\documentclass[11pt,a4paper]{article}
\usepackage[hyperref]{acl2021}
\usepackage{times}
\usepackage{latexsym}

\usepackage{microtype}

\usepackage{caption}
\usepackage{url}
\usepackage{algorithm}
\usepackage{algorithmic}
\usepackage{graphicx}
\usepackage{amsmath}
\usepackage{amsfonts}
\usepackage{multirow}
\usepackage{booktabs}
\usepackage{xcolor}
\usepackage{mathrsfs}
\usepackage{arydshln}
\usepackage{multicol}
\usepackage{multirow}
\usepackage{color}
\usepackage{subfig}
\usepackage{array}
\usepackage{colortbl}
\usepackage{arydshln}
\usepackage{hyperref}
\aclfinalcopy 

\title{Robust Transfer Learning with Pretrained Language Models through Adapters}

\author{Wenjuan Han$^1$\footnotemark[1] $\ $\footnotemark[2], Bo Pang$^2$\footnotemark[1], Yingnian Wu$^2$\\
$^1$ Beijing Institute for General Artificial Intelligence, Beijing, China \\
$^2$ Department of Statistics, University of California, Los Angeles \\
{\tt hanwenjuan@bigai.ai} \\ 
{\tt \{bopang, ywu\}@ucla.edu} \\
\\
}

\date{}

\begin{document}
\maketitle

\renewcommand{\thefootnote}{\fnsymbol{footnote}}
\footnotetext[1]{Equal contributions.}
\footnotetext[2]{Corresponding author.}
\setcounter{footnote}{0}
\renewcommand{\thefootnote}{\arabic{footnote}}

\begin{abstract}
Transfer learning with large pretrained transformer-based language models like BERT has become a dominating approach for most NLP tasks. Simply fine-tuning those large language models on downstream tasks or combining it with task-specific pretraining is often not robust. In particular, the performance considerably varies as the random seed changes or the number of pretraining and/or fine-tuning iterations varies, and the fine-tuned model is vulnerable to adversarial attack. We propose a simple yet effective adapter-based approach to mitigate these issues. Specifically, we insert small bottleneck layers (i.e., adapter) within each layer of a pretrained model, then fix the pretrained layers and train the adapter layers on the downstream task data, with (1) task-specific unsupervised pretraining and then (2) task-specific supervised training (e.g., classification, sequence labeling). Our experiments demonstrate that such a training scheme leads to improved stability and adversarial robustness in transfer learning to various downstream tasks. \footnote{\url{https://github.com/WinnieHAN/Adapter-Robustness.git}}
\end{abstract}


\section{Introduction}

Pretrained transformer-based language models like BERT \citep{devlin-etal-2019-bert}, RoBERTa \citep{liu-etal-2019-robust} have demonstrated impressive performance on various NLP tasks such as sentiment analysis, question answering, text generation, just to name a few. Their successes are achieved through sequential transfer learning \citep{ruder2019neural}: pretrain a language model on large-scale unlabeled data and then fine-tune it on downstream tasks with labeled data. The most commonly used fine-tuning approach is to optimize all parameters of the pretrained model with regard to the downstream-task-specific loss. This training scheme is widely adopted due to its simplicity and flexibility \citep{phang2018sentence,peters2019tune,lan2019albert, raffel2020exploring, clark2020electra, nijkamp-etal-2021-script, lewis2020pre}. 

      \begin{figure}[]
      \begin{center}
      \includegraphics[width=0.9\columnwidth,trim=0 0 0 0,clip]{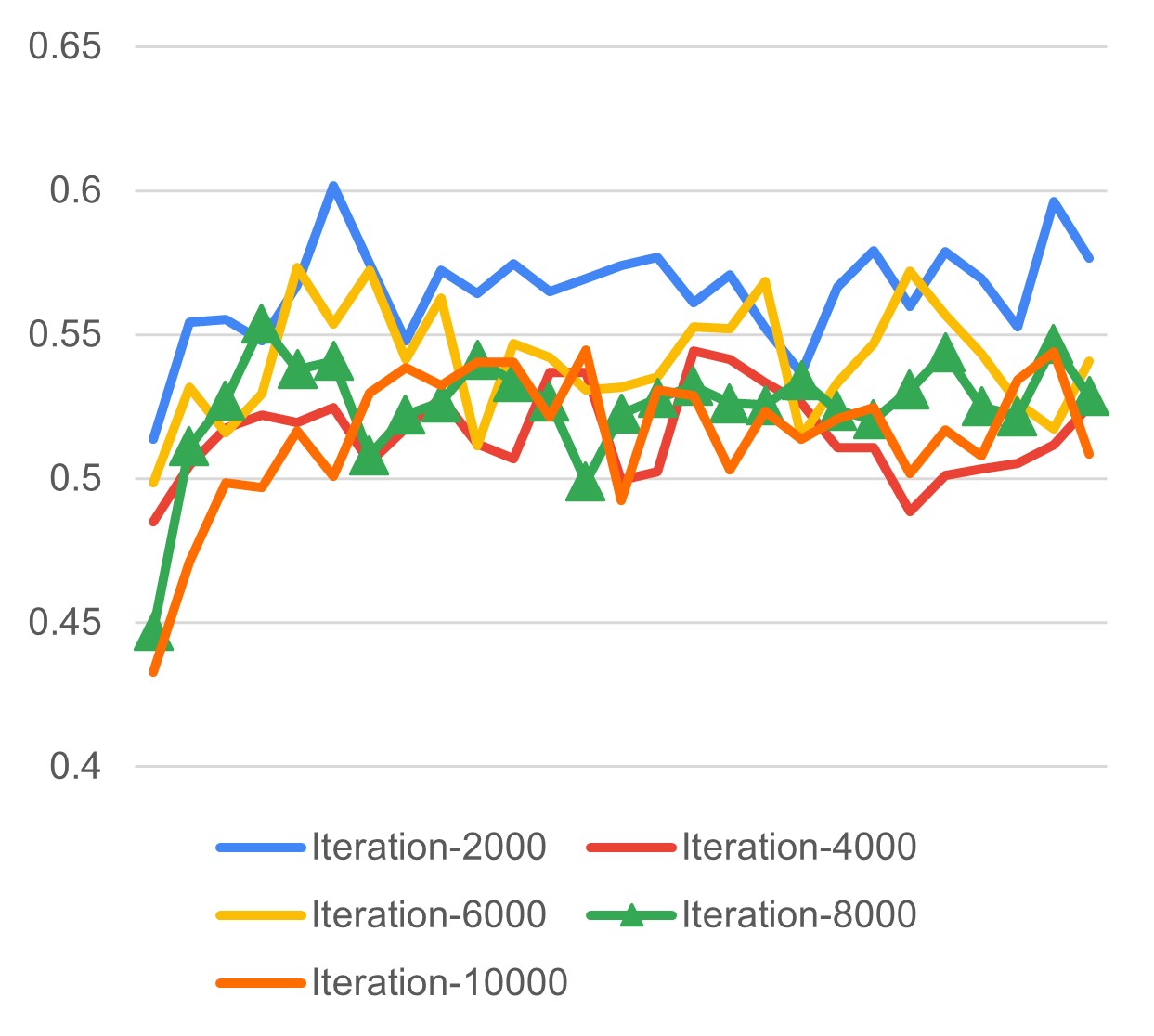}
      \caption{\small Learning curves of fine-tuning with the task-specific pretraining iterations varied. The \textcolor{green}{curve with triangles} represents the model that has converged in the 8000-th pretraining iteration.}
      \label{pic:pretrain-finetune-different-object}
      \vspace{-0.5cm}
      \end{center}
      \end{figure}
      
      \begin{table*}[]
        \centering
        \resizebox{\textwidth}{!}{%
        \begin{tabular}{c|c|c|c|c|c|c|c|c|c|c}
        \hline
        \multicolumn{2}{c|}{}                                     & \textbf{WNLI}     & \textbf{RTE}      & \textbf{MRPC}        & \textbf{STS-B}                 & \textbf{CoLA}           & \textbf{SST-2}    & \textbf{QNLI}     & \textbf{QQP}         & \textbf{MNLI}                          \\ \hline
        \multicolumn{2}{c|}{\textbf{Metrics}}                     & \textit{Acc.} & \textit{Acc.} & \textit{F1/Acc.} & \textit{P/S corr.} & \textit{M corr.} & \textit{Acc.} & \textit{Acc.} & \textit{Acc./F1} & \textit{M acc.} \\ \hline
        \multicolumn{2}{c|}{\textbf{WO.}}                         & 56.34             & 65.7              & 88.85/84.07          & 88.64/88.48                    & 56.53                   & 92.32             & 90.66             & 90.71/87.49          & 84.10                            \\ \hline
        \multirow{2}{*}{\textbf{W.}} & \textit{F.}          & 45.07             & 61.73             & 89.47/85.29          & 83.95/83.70                    & 49.23                   & 91.97             & 87.46             & 88.40/84.31          & 81.08                            \\ \cline{2-11} 
                                     & \textit{TSP.+F.} & 56.34             & 68.59             & 89.76/86.37          & 89.24/88.87                    & 64.87                   & 92.78             & 91.12             & 90.92/87.88          & 84.14                                \\ \hline
        \end{tabular}%
        }
        \caption{Performance on the development dataset of GLUE. Results of W.(F.) are reported in \href{https://github.com/Adapter-Hub/adapter-transformers/tree/master/examples/text-classification}{Adapter-Hub}. We report results of WO. using the implementation from \citet{wolf-etal-2020-transformers}. Acc.: Accuracy. M acc.: Mismatched Acc. P/S acc.: Person/Spearman corr. M corr.: Matthew's corr. TSP.: Task-Specific Pretrain. F.: Finetune. WO.: Without adapter. W.: With adapter.}
        \label{tab:glue_table}
        \end{table*}

Despite the success of the standard sequential transfer learning approach, recent works \citep{gururangan2020don,lee2020biobert,bertweet} have explored domain-specific or task-specific unsupervised pretraining, that is, masked language model training on the downstream task data before the final supervised fine-tuning on it. And they demonstrated benefits of task-specific pretraining on transfer learning performance. However, both standard sequential transfer learning and that with task-specific pretraining are unstable in the sense that downstream task performance is subject to considerable fluctuation while the random seed is changed or the number of pretraining and/or fine-tuning iterations is varied even after the training has converged (see Section~\ref{sec:seed} and Section~\ref{sec:iter} for details).
Inspired by \citet{he2021effectiveness}, analyzing the instability w.r.t different random seeds, we analyze other aspects of the stability: the stability to pretraining and fine-tuning iterations and the gradient vanishing issue.
For instance, as observed in Fig.~\ref{pic:pretrain-finetune-different-object}, as the number of task-specific pretraining iteration varies, CoLA’s performance is severely unstable in fine-tuning. Besides instability, we also observe that task-specific pretraining is vulnerable to adversarial attack. Last but not least, task-specific pretraining and/or fine-tuning on the entire model is highly parameter-inefficient given the large size of these models (e.g., the smallest BERT has $110$ million parameters). 


In this work, we propose a simple yet effective adapter-based approach to mitigate these issues. Adapters are some small bottleneck layers inserted within each layer of a pretrained model \citep{houlsby2019parameter, pfeiffer2020adapterfusion, pfeiffer2020adapterhub}. The adapter layers are much smaller than the pretrained model in terms of the number of parameters. For instance, the adapter used in \citet{houlsby2019parameter} only adds $3.6\%$ parameters per task. In our approach, we adapt the pretrained model to a downstream task through 1) task-specific pretraining and 2) task-specific supervised training (namely, fine-tuning) on the downstream task (e.g., classification, sequence labeling) by only optimizing the adapters and keeping all other layers fixed. Our approach is parameter-efficient given that only a small number of parameters are learned in the adaptation. 

The adapted model learned through our approach can be viewed as a residual form of the original pretrained model. Suppose $x$ is an input sequence and $h_{\rm original}$ is the features of $x$ computed by the original model. Then the feature computed by the adapted model is,
\begin{equation}
h_{\rm adapted} = h_{\rm original} + f_{\rm adapter}(x),    
\end{equation}
where $f_{\rm adapter}(x)$ is the residual feature in addition to $h_{\rm original}$ and $f_{\rm adapter}$ is the adapter learned in the adaptation process. $h_{\rm original}$ extracts general features that are shared across tasks, while $f_{\rm adapter}$ is learned to extract task-specific features. In prior work \citep{houlsby2019parameter, pfeiffer2020adapterhub},  $f_{\rm adapter}$ is learned with task-specific supervised learning objective, distinctive from the unsupervised pretraining objective, and might not be compatible with $h_{\rm original}$, as evidenced in our experiments. In our approach, $f_{\rm adapter}$ is first trained with the same pretraining objective\footnote{In this work, we conduct experiments with the most widely used pretraining objective, masked language modeling. The same training scheme can be extended to other pretraining objectives.} on the task-specific data before being adapted with the supervised training objective, encouraging the compatibility between $h_{\rm original}$ and $f_{\rm adapter}$, which is shown to improve the downstream task performance in our experiments (see Table~\ref{tab:glue_table}).

Some prior works have examined the potential causes of the instability of pretrained language models in transfer learning. \citet{lee2019mixout} proposed that catastrophic forgetting in sequential transfer learning underlined the instability, while \citet{mosbach2020stability} proposed that gradient vanishing in fine-tuning caused it. Pinpointing the cause of transfer learning instability is not the focus of the current work, but our proposed method seems to able to enhance transfer learning on both aspects.

The standard sequential transfer learning or that with task-specific pretraining updates all model parameters in fine-tuning. In contrast, our approach keeps the pretrained parameters unchanged and only updates the parameters in the adapter layers, which are a small amount compared to the pretrained parameters. Therefore, our approach naturally alleviates catastrophic forgetting considering the close distance between the original pretrained model and the adapted model. On the other hand, we do not observe gradient vanishing with our transfer learning scheme (see Section~\ref{sec:seed} for more details). This might be because optimizing over a much smaller parameter space in our approach, compared to the standard sequential transfer learning scheme where all parameters are trained, renders the optimization easier. We leave it to future work for further theoretical analysis. 

In addition to its improved stability, the proposed transfer learning scheme is also likely to be more robust to adversarial attack. Given that it updates the entire model, the standard transfer learning approach might suffer from overfitting to the downstream task, and thus a small perturbation in the input might result in consequential change in the model prediction. In turn, it might be susceptible to adversarial attack. Our approach only updates a much smaller portion of parameters, and hence might be more robust to these attacks, which is confirmed in our empirical analysis (see Section~\ref{sec: adv}). 



\textbf{Contributions.} In summary our work has the following contributions. (1) We propose a simple and parameter-efficient approach for transfer learning. (2) We demonstrate that our approach improves the stability of the adaptation training and adversarial robustness in downstream tasks. (3) We show the improved performance of our approach over strong baselines. Our source code is publicly available at \url{https://github.com/WinnieHAN/Adapter-Robustness.git}.

      \begin{figure}[]
      \begin{center}
      \includegraphics[width=0.7\columnwidth,trim=0 0 0 2,clip]{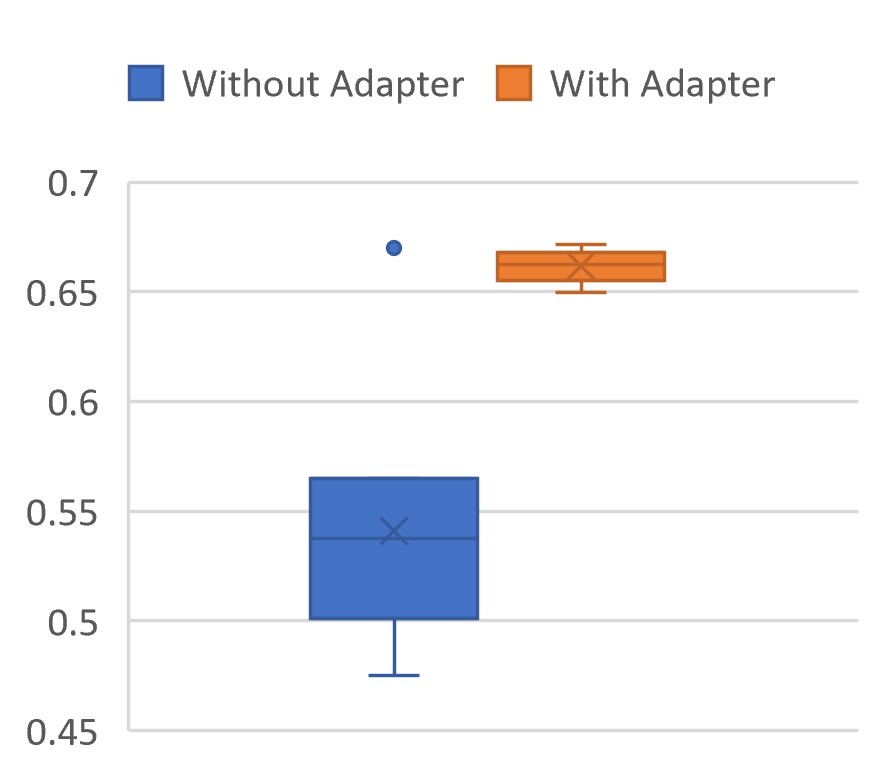}
      \caption{\small Distribution of dev scores on RTE from 10 random seed restarts when finetuning (1) BERT \citep{devlin-etal-2019-bert} and (2) BERT with the adapter architecture.}
      \label{pic:randomseed}
      \vspace{-0.5cm}
      \end{center}
      \end{figure}

      \begin{figure}[]
      \begin{center}
      \includegraphics[width=0.7\columnwidth,trim=0 0 0 2,clip]{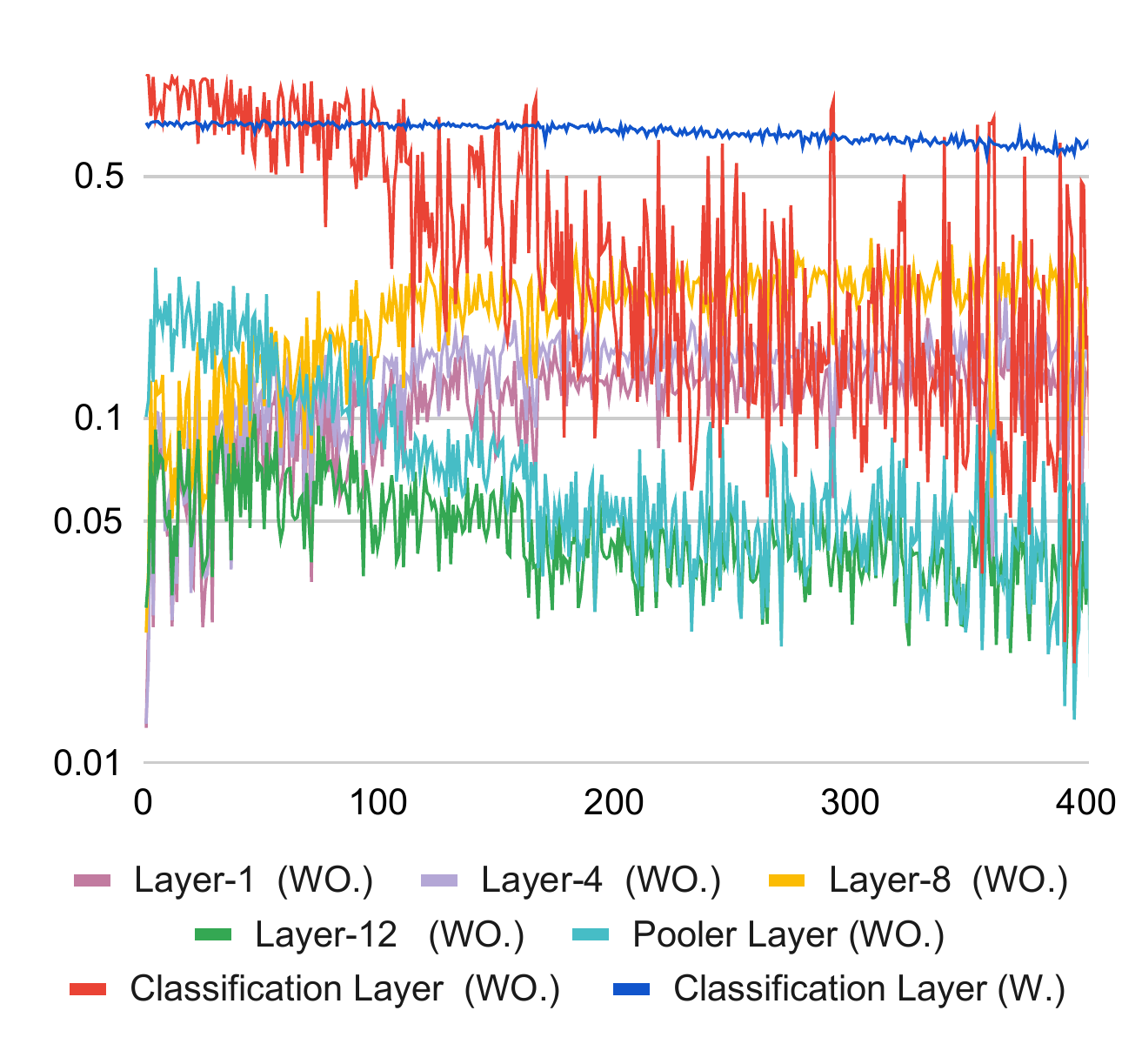}
      \caption{\small Gradient norms (on log scale) of intermediate layer and classification layer on RTE for with/without-adapter fine-tuning run. WO.: Without adapter. W.: With adapter.}
      \label{pic:gv}
      \vspace{-0.5cm}
      \end{center}
      \end{figure}
    
      \begin{figure}[]
      \begin{center}
      \includegraphics[width=0.7\columnwidth,trim=0 0 0 2,clip]{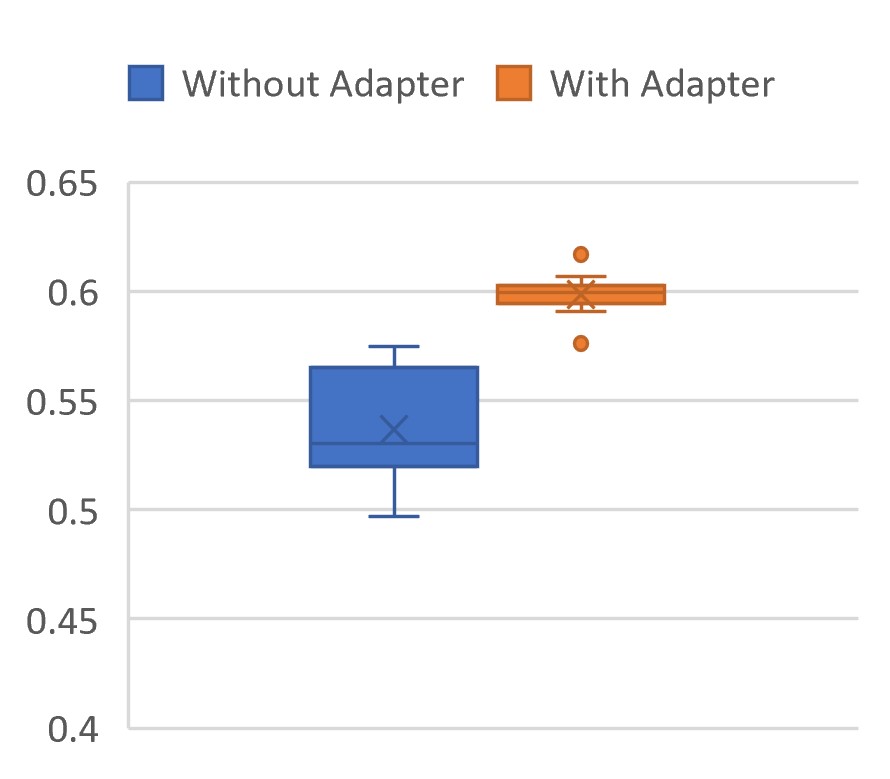}
      \caption{\small Box plots showing the TSP. stability of BERT with/without adapter on CoLA.}
      \label{pic:Pretraining-Iterations}
      \vspace{-0.5cm}
      \end{center}
      \end{figure}

      \begin{figure}[]
      \begin{center}
      \includegraphics[width=0.7\columnwidth,trim=0 0 0 2,clip]{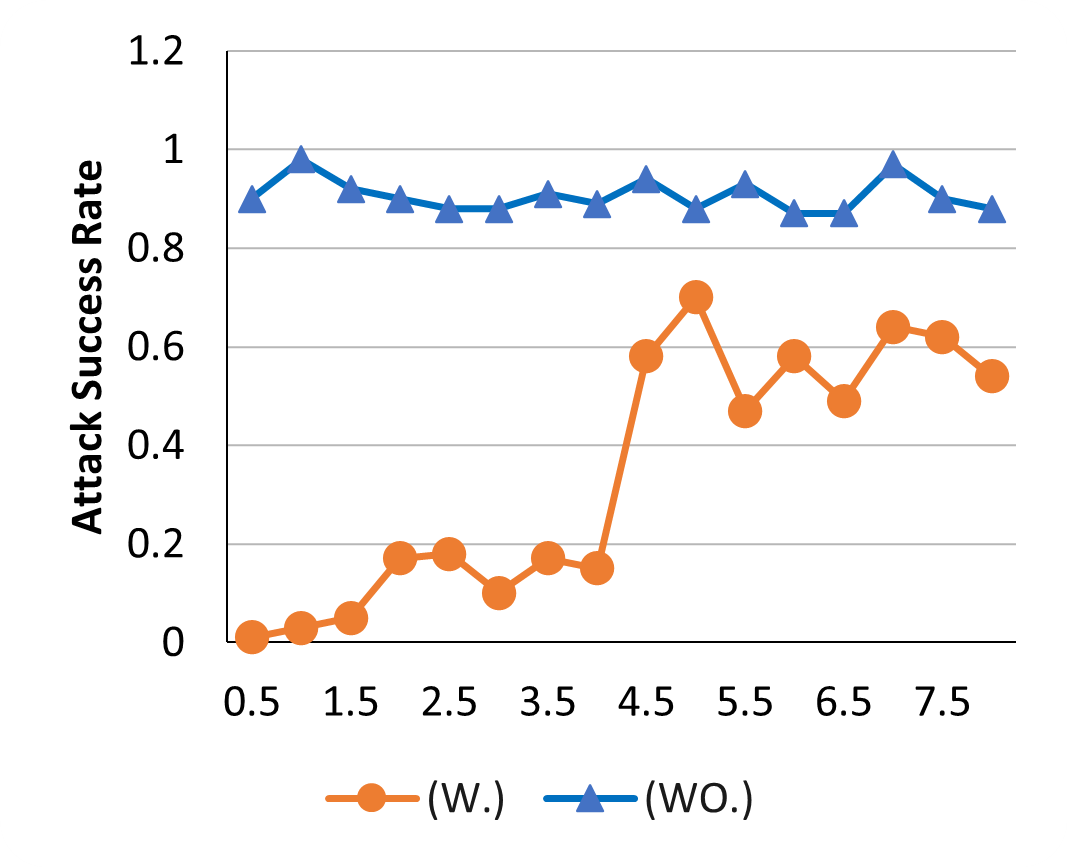}
      \caption{\small Attack success rate of BERT with/without adapter during task-specific pretraining. WO.: Without adapter. W.: With adapter.}
      \label{pic:attack_success_rate}
      \end{center}
      \end{figure}

    \begin{figure*}[]
    \centering
    \begin{minipage}[]{0.32\textwidth} 
    \centering
    \includegraphics[width=0.95\textwidth]{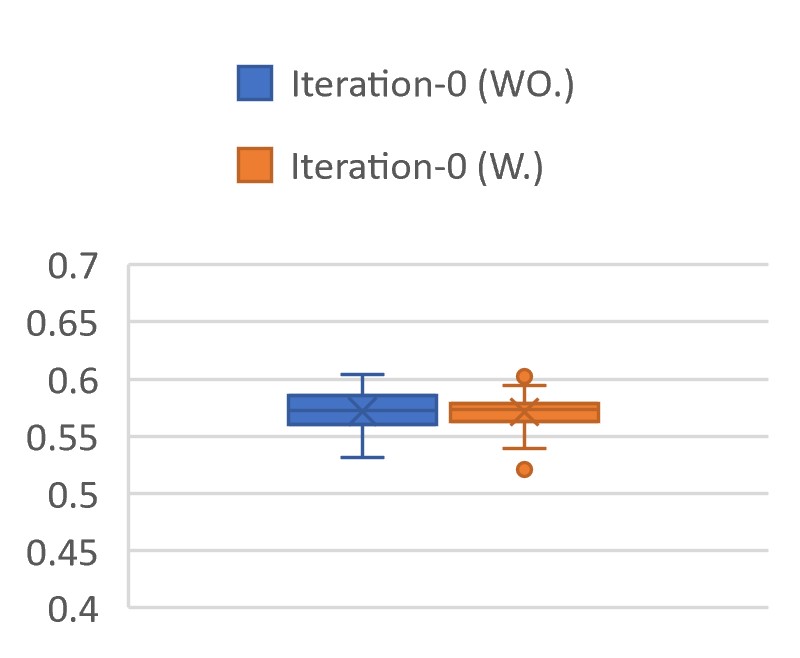}
    \captionsetup{labelformat=empty}
    \caption{(a) Pretraining Iteration 0}
    \end{minipage}
    \begin{minipage}[]{0.32\textwidth}
    \centering
    \includegraphics[width=0.95\textwidth]{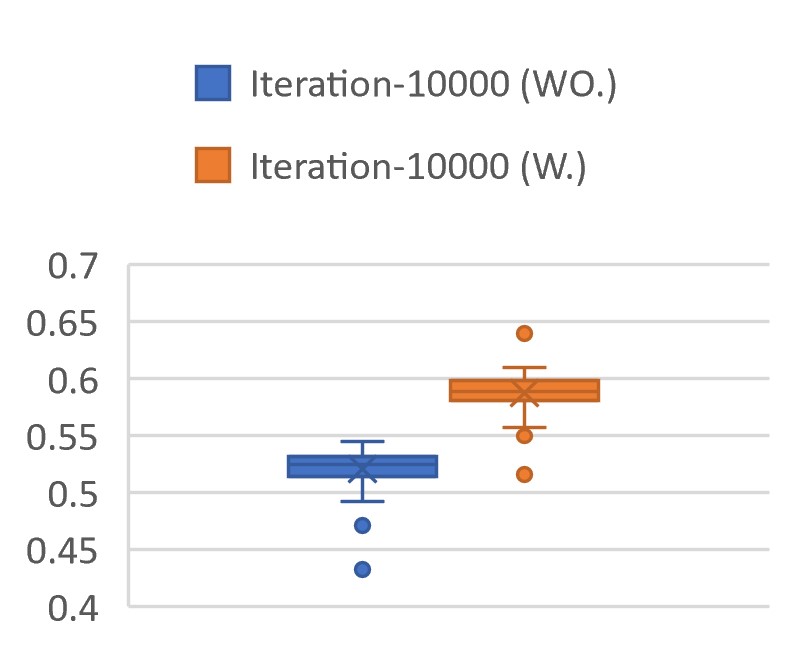}
    \captionsetup{labelformat=empty}
    \caption{(b) Pretraining Iteration 10000}
    \end{minipage}
    \begin{minipage}[]{0.32\textwidth}
    \centering
    \includegraphics[width=0.95\textwidth]{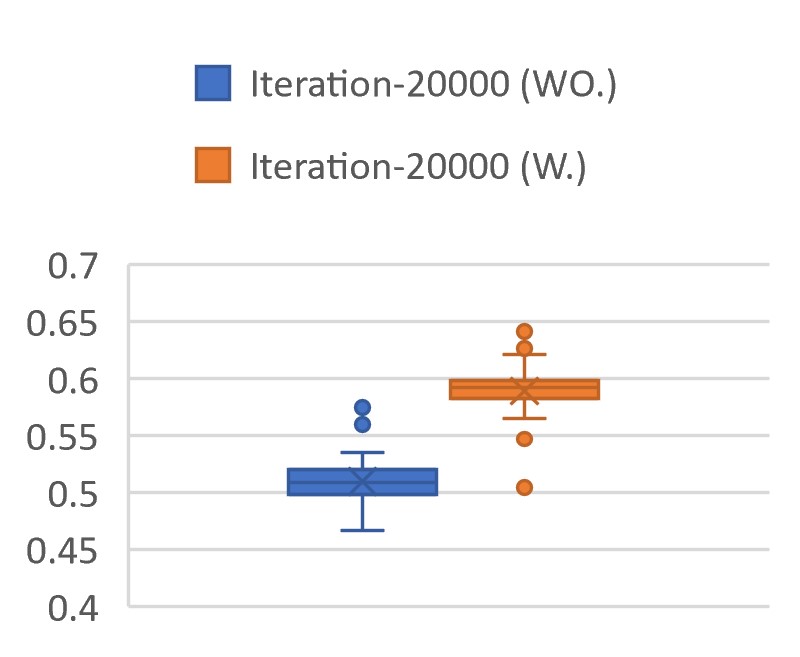}
    \captionsetup{labelformat=empty}
    \caption{(c) Pretraining Iteration 20000}
    \end{minipage}
    \caption{\small Box plots showing the fine-tuning stability of BERT with/without adapter for different TSP. iterations on CoLA. WO.: Without adapter. W.: With adapter.}
    \label{pic:Finetune-Iterations}
    \end{figure*}

\section{Instability to Different Random Seeds}
\label{sec:seed}

We first evaluate the training instability with respect to multiple random seeds: fine-tuning the model multiple times in the same setting, varying only the random seed. We conduct the experiments on RTE \citep{wang2018glue} when fine-tuning 1) BERT-base-uncased \citep{devlin-etal-2019-bert} and 2) BERT-base-uncased with the adapter \citep{houlsby2019parameter} \footnote{For all the experiments, we use the implementation of \citet{pfeiffer2020adapterhub}: \url{https://github.com/Adapter-Hub/adapter-transformers.git}.}. As shown in Figure \ref{pic:randomseed}, the model without adapter leads to a large standard deviation on the fine-tuning accuracy, while the one with adapter results in a much smaller variance on the task performance.

\paragraph{Gradient Vanishing}
\citet{mosbach2020stability} argues that the fine-tuning instability can be explained by optimization difficulty and gradient vanishing. In order to inspect if the adapter-based approach suffers from this optimization problem, we plot the $L_2$ gradient norm with respect to different layers of BERT, pooler layer and classification layer, for fine-tuning with or without adapter in Figure \ref{pic:gv}. 

In traditional fine-tuning (without adapter), we see vanishing gradients for not only the top layers but also the pooler layer and classification layer. This is in large contrast to the with-adapter fine-tuning. The gradient norm in the with-adapter fine-tuning does not decrease significantly in the training process. These results imply that the adaptation with adapter does not exhibit gradient vanishing and presents a less difficult optimization problem, which in turn might explain the improved stability of our approach.

\section{Instability to Pretraining and Fine-tuning Iterations}
\label{sec:iter}

Fine-tuning with all parameters also exhibits another instability issue. In particular, fine-tuning a model multiple times on the pretrained language model, varying the task-specific pretraining iterations and fine-tuning iterations, leads to a large standard deviation in downstream task performance. As observed in Figure \ref{pic:pretrain-finetune-different-object}, CoLA’s performance when varying the task-specific pretraining iterations is severely unstable during pretraining iterations and fine-tuning iterations. The model has converged at the pretraining iteration of 8000. However, fine-tuning based on this model does not obtain the best performance.

\paragraph{Pretraining Iterations.}
Figure \ref{pic:Pretraining-Iterations} displays the performance on CoLA of 10 fine-tuning runs with and without the adapter. For each run, we vary only the number of pretraining iterations from 2000 to 20000 with an interval of 2000 and fix the fine-tuning epochs to 10.
We clearly observe that most runs for BERT with adapter outperforms the one without adapter. Moreover, the adapter makes pretraining BERT significantly more stable than the standard approach (without adapter).

\paragraph{Fine-tuning Iterations.}
We then study the stability with regard to the number of fine-tuning iterations.
We show box plots for BERT using various pretraining iterations and fine-tuning iterations, with and without adapter in Figure~\ref{pic:Finetune-Iterations}.
The three sub-figures represent the early, mid, and late stages of pretraining, corresponding to the $0$-th, $10000$-th, and $20000$-th iteration respectively. The $0$-th iteration represents the original model without task-specific pretraining. The model suffers from underfitting in the $0$-th iteration and overfitting in the $20000$-th iteration.

In Figure~\ref{pic:Finetune-Iterations} (a), we plot the distributions of the development scores from 100 runs when fine-tuning BERT with various fine-tuning epochs ranging from 1 to 100. In the early stage, the average development score of the model with the adapter is a little lower than the baseline model while the stability is better. After several epochs of pretraining, the adapter gradually shows improved performance in terms of the mean, minimum and maximum as demonstrated in Figure~\ref{pic:Finetune-Iterations} (b). In the end of the pretrainig, there exists an over-fitting problem for the traditional BERT models. Pretraining transfers the model to a specific domain and fails to maintain the original knowledge. In contrast, the performance with the adapter still grows as training continues and consistently benefit from pretraining.  Besides, we observe that the adapter leads to a small variance in the fine-tuning performance, especially in the late stage. Additional plots and learning curves can be found in the Appendix.

\section{Adversarial Robustness}
\label{sec: adv}

While successfully applied to many domains, the predictions of Transformers \citep{vaswani2017attention} become unreliable in the presence of small adversarial perturbations to the input \citep{sun2020adv,li2020bert}. 
Therefore, the adversarial attacker has become an important tool \citep{moosavi2016deepfool} to verify the robustness of models.
The robustness is usually evaluated from attack effectiveness (i.e., attack success rate).
We use a SOTA adversarial attack approach to assess the robustness: PWWS attacker \citep{ren2019generating}. \footnote{We use the implementation in OpenAttack toolkit \url{https://github.com/thunlp/OpenAttack.git}. It generates adversarial examples and evaluation the adversarial robustness of the victime model using thee adversarial examples. We use the default settings including all the hyper-parameter values.}. 
Figure \ref{pic:attack_success_rate} shows the attack success rate of BERT with/without adapter during task-specific pretraining on SST-2. 
The x-axis is the number of epochs for task-specific pretraining. It can be observed that the model with the adapter has better adversarial robustness.


\section{Conclusion}
We propose a simple yet effective transfer learning scheme for large-scale pretrained language model. We insert small bottleneck layers (i.e., adapter) within each block of the pretrained model and then optimize the adapter layers in task-specific unsupervised pretraining and supervised training (i.e., fine-tuning) while fixing the pretrained layers. Extensive experiments demonstrate that our approach leads to improved stability with respect to different random seeds and different number of iterations in task-specific pretraining and fine-tuning, enhanced adversarial robustness, and better transfer learning task performance. We therefore consider the proposed training scheme as a robust and parameter-efficient transfer learning approach.


\section*{Acknowledgments}
Y. W. is partially supported by NSF DMS 2015577.

\bibliographystyle{acl_natbib}
\bibliography{anthology,acl2021}

\newpage 
\appendix
\section{Hyper-Parameters Setting}
We conduct the experiments on the task of GLUE tasks \citep{wang2018glue} when finetuning 1) BERT-base-uncased \citep{devlin-etal-2019-bert} and 2) BERT-base-uncased with the adapter architecture \citep{houlsby2019parameter}. For all the experiments, we use the implementation from \url{https://github.com/Adapter-Hub/adapter-transformers.git}. For the model with adapter, we follows the setup from \citet{mosbach2020stability}. For all experiments, we use the default hyper-parameters except for the number of epochs. Please refer to the provided link.


The main hyper-parameters are listed in Table \ref{tab:parameters-w} and Table \ref{tab:parameters-wo}.
\begin{table}[H]
\centering
\begin{tabular}{ll}
\hline
Max Sequence Length          &    256    \\
Batch Size & 32  \\
Learning rate      &    1e-4    \\
Number of Epochs                       & 20  \\
\hline
\end{tabular}%
\caption{Hyper-parameters for BERT with Adapter.}
\label{tab:parameters-w}
\end{table}

\begin{table}[H]
\centering
\begin{tabular}{ll}
\hline
Max Sequence Length          &    128    \\
Batch Size & 32  \\
Learning rate      &    2e-5    \\
Number of Epochs                       & 10  \\
\hline
\end{tabular}%
\caption{Hyper-parameters for BERT without Adapter.}
\label{tab:parameters-wo}
\end{table}

\section{Instability to Pretraining and Fine-tuning Iterations}
\begin{figure*}[]
      \begin{center}
      \includegraphics[width=2.0\columnwidth,trim=0 0 0 2,clip]{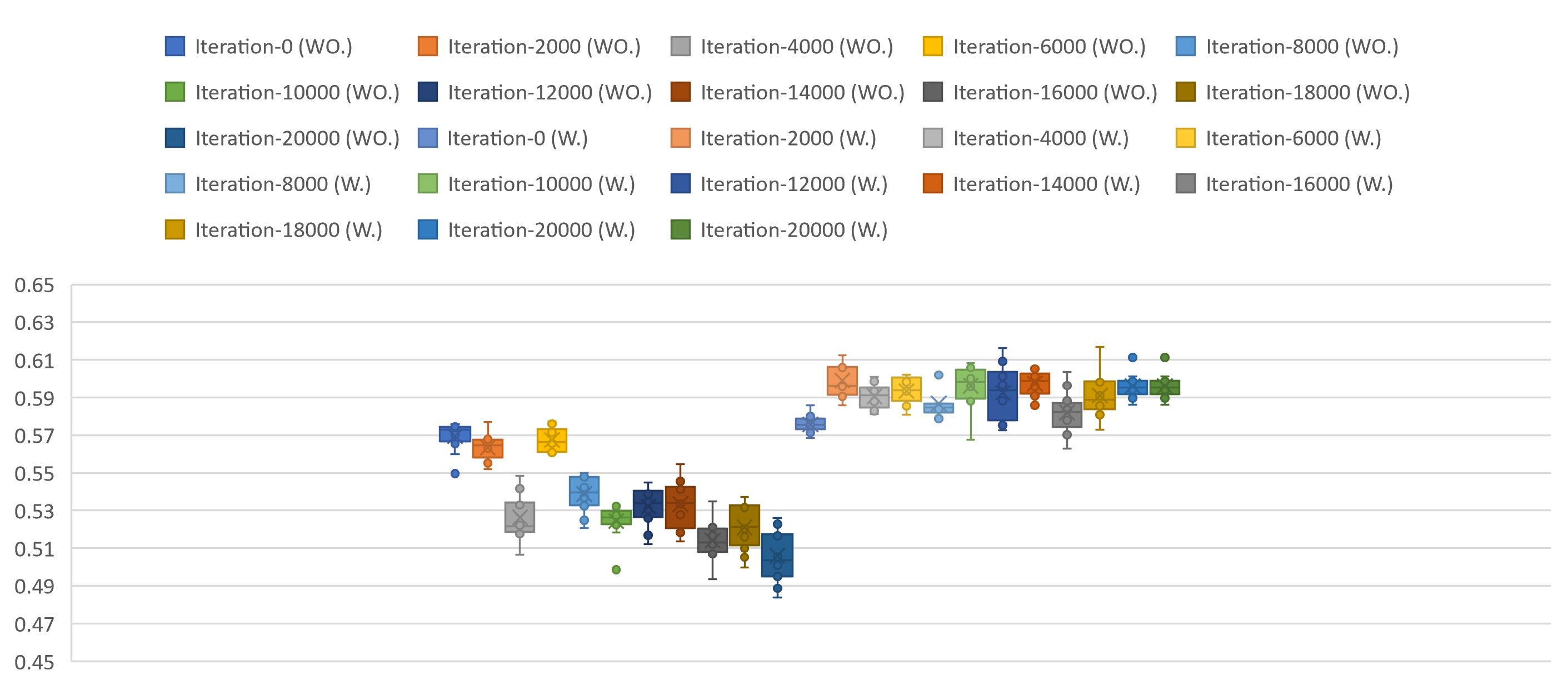}
      \caption{Box plots showing the fine-tuning stability of BERT with/without adapter for different pretraining iteration from 0 to 20000.}
      \label{pic:all}
      \end{center}
      \end{figure*}

\begin{figure*}[]
\centering
\begin{minipage}[]{0.48\textwidth} 
\centering
\includegraphics[width=0.95\textwidth]{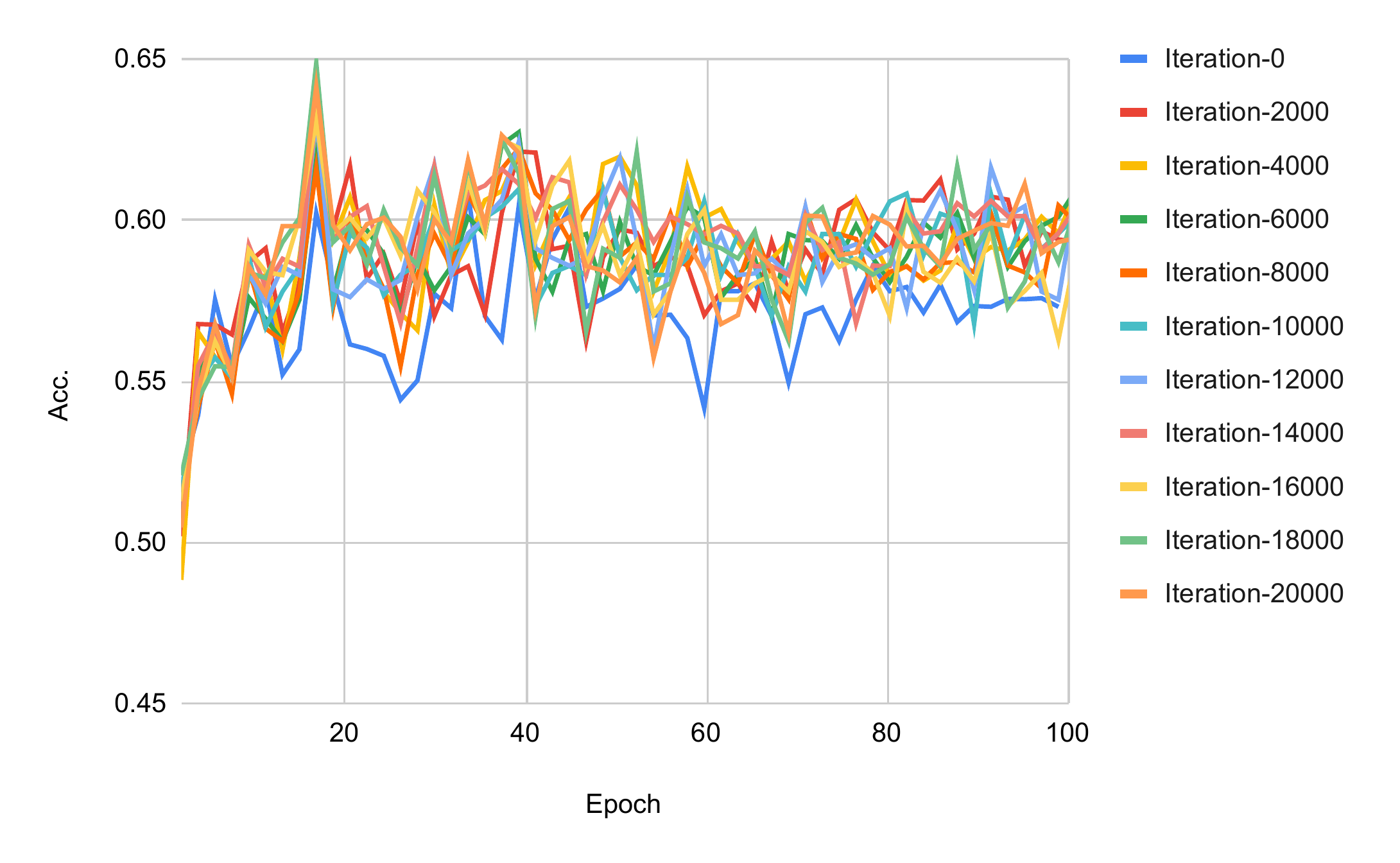}
\captionsetup{labelformat=empty}
\caption{(a) BERT with adapter.}
\end{minipage}
\begin{minipage}[]{0.48\textwidth}
\centering
\includegraphics[width=0.95\textwidth]{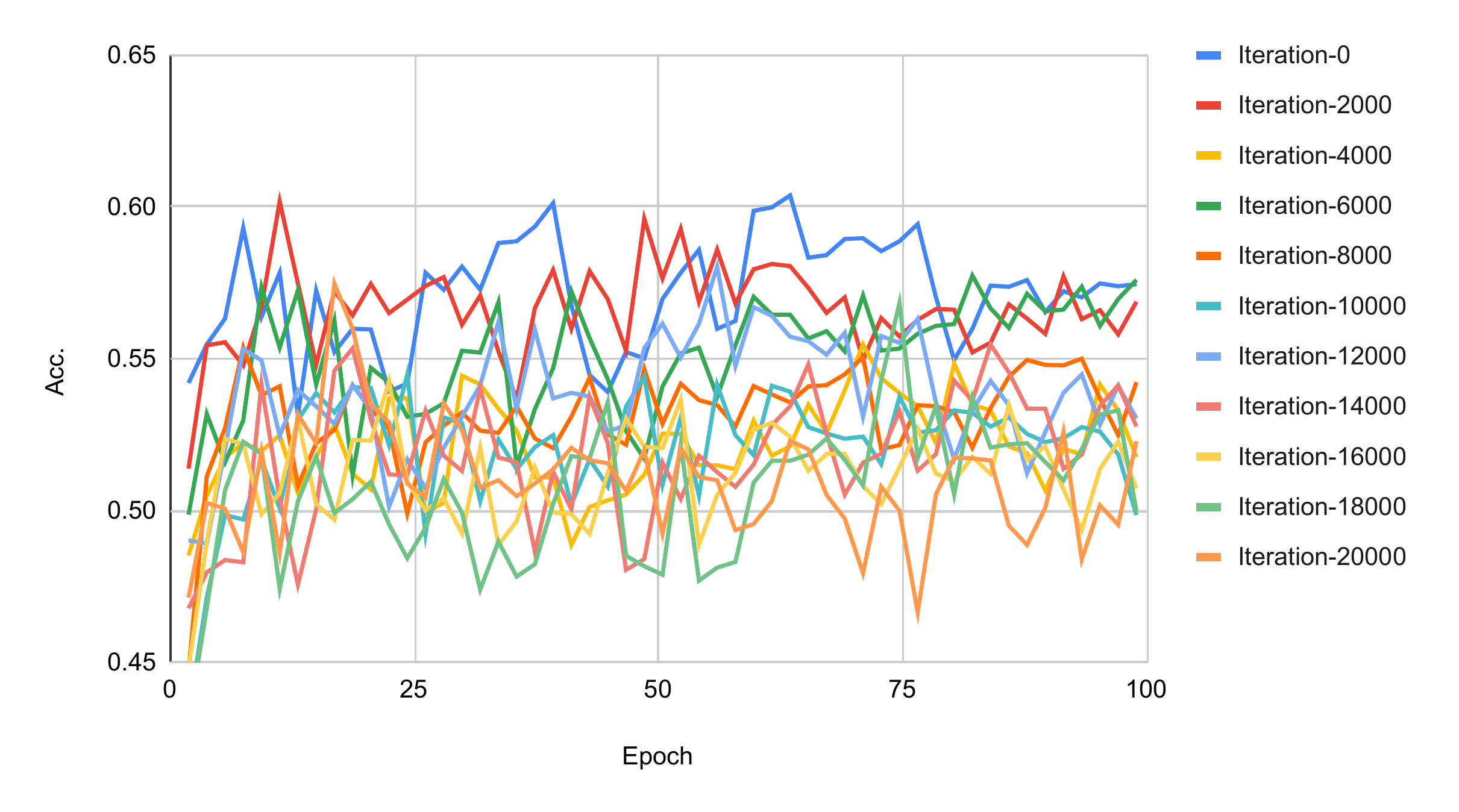}
\captionsetup{labelformat=empty}
\caption{(b) BERT without adapter.}
\end{minipage}
\caption{Learning curves of fine-tuning when varying the pretraining iterations.}
\label{pic:learning_curves_wo}
\end{figure*}

We provide box plots for BERT using various pretraining iterations and fine-tuning iterations, with and without adapter on CoLA in Figure \ref{pic:all}. The corresponding learning curves are in Figure \ref{pic:learning_curves_wo}.

\section{Instability for Large Dataset}
In contrast to relatively large datasets, smaller data is more suitable and convincing as an example to analyze stability. Small dataset is easier to encounter over-fitting problems and often not stable \cite{devlin-etal-2019-bert}. We use MNLI to evaluate the training instability in terms of 5 random seeds with the same setup in Figure \ref{pic:randomseed}. The interquartile range of BERT with adapter on the distribution of dev scores is smaller than BERT without adapter. It shows that the model without adapter consistently leads to the instability issue on the fine-tuning accuracy, while the adapter architecture brings less benefit with larger dataset.

\end{document}